\documentclass[11pt,a4paper]{article}
\usepackage{times,latexsym}
\usepackage{url}
\usepackage[T1]{fontenc}
\usepackage{totcount}

\usepackage[acceptedWithA]{tacl2018v2}

\usepackage{xspace,mfirstuc,tabulary}

\newif\iftaclinstructions
\taclinstructionsfalse 
\iftaclinstructions
\renewcommand{\confidential}{}
\renewcommand{\anonsubtext}{(No author info supplied here, for consistency with
TACL-submission anonymization requirements)}
\newcommand{\instr}
\fi

\iftaclpubformat 

\else

\fi

\usepackage{bbding}
\usepackage{pifont}
\usepackage{graphicx}

\usepackage{booktabs}
\usepackage{multirow}
\usepackage{array}
\usepackage[export]{adjustbox}
\usepackage{float}
\usepackage{xcolor}

\usepackage{microtype}

\newenvironment{itemize*}%
  {\begin{itemize}%
    \setlength{\itemsep}{0.9pt}%
    \setlength{\parskip}{0.9pt}%
    \setlength{\topsep}{0.9pt}}%
  {\end{itemize}}

\newtotcounter{citnum} 
\def\oldbibitem{} \let\oldbibitem=\bibitem
\def\bibitem{\stepcounter{citnum}\oldbibitem}

\title{A Primer in BERTology: What We Know About How BERT Works}

\author{Anna Rogers \\
  \small{Center for Social Data Science}\\
  \small{University of Copenhagen}\\
  \texttt{\small{arogers@sodas.ku.dk}}\\\And
  Olga Kovaleva \\
  \small{Dept. of Computer Science}\\  
  \small{University of Massachusetts Lowell}\\
  \texttt{\small{okovalev@cs.uml.edu}}\\\And
  Anna Rumshisky \\
  \small{Dept. of Computer Science}\\    
 \small{University of Massachusetts Lowell}\\
 \texttt{\small{arum@cs.uml.edu}}}

\date{}

\begin{document}
\maketitle

\begin{abstract}

Transformer-based models have pushed state of the art in many areas of NLP, but our understanding of what is behind their success is still limited. This paper is the first survey of over 150 studies of the popular BERT model. We review the current state of knowledge about how BERT works, what kind of information it learns and how it is represented, common modifications to its training objectives and architecture, the overparameterization issue and approaches to compression. We then outline directions for future research.

\end{abstract}

\section{Introduction}

Since their introduction in 2017, Transformers \cite{vaswani2017attention} have taken NLP by storm, offering enhanced parallelization and better modeling of long-range dependencies. The best known Transformer-based model is BERT \cite{devlin2019bert}; it obtained state-of-the-art results in numerous benchmarks and is still a must-have baseline. 

While it is clear that BERT works remarkably well, it is less clear \textit{why}, which limits further hypothesis-driven improvement of the architecture. Unlike CNNs, the Transformers have little cognitive motivation, and the size of these models limits our ability to experiment with pre-training and perform ablation studies. 
This explains a large number of studies over the past year that attempted to understand the reasons behind BERT's performance.

In this paper, we provide an overview of what has been learned to date, highlighting the questions which are still unresolved. We first consider the linguistic aspects of it, i.e., the current evidence regarding the types of linguistic and world knowledge learned by BERT, as well as where and how this knowledge may be stored in the model. We then turn to the technical aspects of the model and provide an overview of the current proposals to improve BERT's architecture, pre-training and fine-tuning. We conclude by discussing the issue of overparameterization, the approaches to compressing BERT, and the nascent area of pruning as a model analysis technique.

\section{Overview of BERT architecture}
\label{sec:bert}

Fundamentally, BERT is a stack of Transformer encoder layers  \cite{vaswani2017attention} which consist of multiple self-attention "heads". For every input token in a sequence, each head computes key, value and query vectors, used to create a weighted representation. The outputs of all heads in the same layer are combined and run through a fully-connected layer. Each layer is wrapped with a skip connection and followed by layer normalization.

The conventional workflow for BERT consists of two stages: pre-training and fine-tuning. Pre-training uses two self-supervised tasks: masked language modeling (MLM, prediction of randomly masked input tokens) and next sentence prediction (NSP, predicting if two input sentences are adjacent to each other). In fine-tuning for downstream applications, one or more fully-connected layers are typically added on top of the final encoder layer.

The input representations are computed as follows: each word in the input is first tokenized into wordpieces \cite{wu2016google}, and then three embedding layers (token, position, and segment) are combined to obtain a fixed-length vector. Special token \texttt{[CLS]} is used for classification predictions, and \texttt{[SEP]} separates input segments. 

Google\footnote{\url{https://github.com/google-research/bert}} and HuggingFace
\cite{WolfDebutEtAl_2020_HuggingFaces_Transformers_State-of-the-art_Natural_Language_Processing} provide many variants of BERT, including the original "base" and "large" versions. They vary in the number of heads, layers, and hidden state size.

\section{What knowledge does BERT have?}
\label{sec:knowledge}

A number of studies have looked at the knowledge encoded in BERT weights. The popular approaches include fill-in-the-gap probes of MLM, analysis of self-attention weights, and probing classifiers with different BERT representations as inputs.

\subsection{Syntactic knowledge}
\label{sec:syntax}

\citet{lin2019open} showed that \textbf{BERT representations are hierarchical rather than linear}, i.e. there is something akin to syntactic tree structure in addition to the word order information. \citet{TenneyXiaEtAl_2019_What_do_you_learn_from_context_Probing_for_sentence_structure_in_contextualized_word_representations} and \citet{liu2019linguistic} also showed that \textbf{BERT embeddings encode information about parts of speech, syntactic chunks and roles}. 
Enough syntactic information seems to be captured in the token embeddings themselves
to recover syntactic trees \cite{vilares2020parsing,kim2020pre,rosa2019inducing}, although probing classifiers could not recover the labels of distant parent nodes in the syntactic tree 
\cite{liu2019linguistic}. \citet{WarstadtBowman_2020_Can_neural_networks_acquire_structural_bias_from_raw_linguistic_data} report evidence of hierarchical structure in three out of four probing tasks.

As far as \textit{how} syntax is represented, it seems that \textbf{syntactic structure is not directly encoded in self-attention weights}. \citet{htut2019attention} were unable to extract full parse trees from BERT heads even with the gold annotations for the root. \citet{jawahar2019does} include a brief illustration of a dependency tree extracted directly from self-attention weights, but provide no quantitative evaluation. 

However, \textbf{syntactic information can be recovered from BERT token representations}. \citet{hewitt2019structural} were able to learn transformation matrices that successfully recovered syntactic dependencies in PennTreebank data from BERT's token embeddings \cite[see also][]{ManningClarkEtAl_2020_Emergent_linguistic_structure_in_artificial_neural_networks_trained_by_self-supervision}. 
\citet{jawahar2019does} experimented with transformations of the [CLS] token using Tensor Product Decomposition Networks  \cite{McCoyLinzenEtAl_2019_RNNs_implicitly_implement_tensor-product_representations}, concluding that dependency trees are the best match among 5 decomposition schemes (although the reported MSE differences are very small). \citet{MiaschiDellOrletta_2020_Contextual_and_Non-Contextual_Word_Embeddings_in-depth_Linguistic_Investigation} performs a range of syntactic probing experiments with concatenated token representations as input. 

\begin{figure}[!t]
    \centering
    \includegraphics[width=.9\columnwidth]{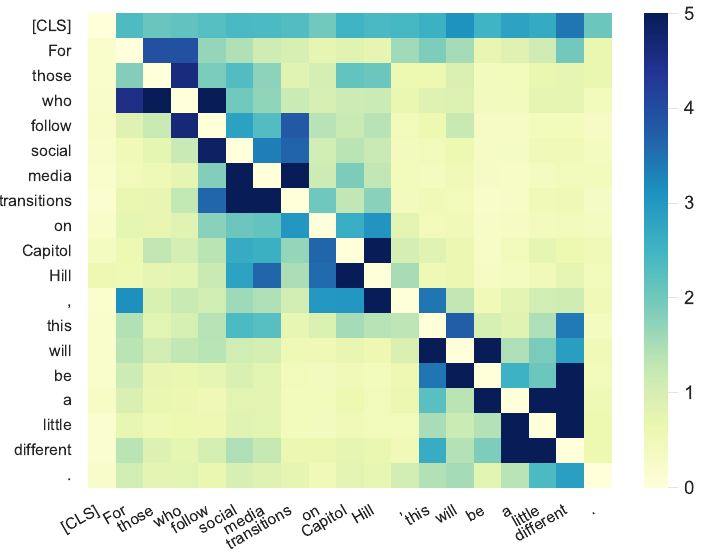}
    \includegraphics[width=.8\columnwidth]{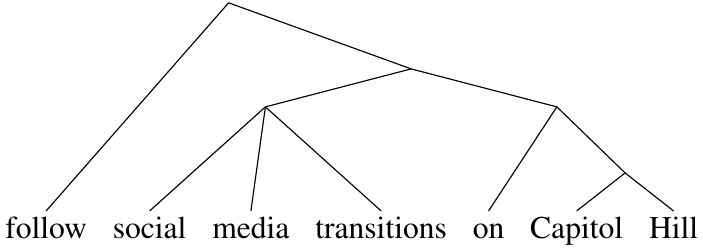}    
    \caption{Parameter-free probe for syntactic knowledge: words sharing syntactic subtrees have larger impact on each other in the MLM prediction     \cite{WuChenEtAl_2020_Perturbed_Masking_Parameter-free_Probing_for_Analyzing_and_Interpreting_BERT}}
    \label{fig:wu-syntax}
\end{figure}

Note that all these approaches look for the evidence of gold-standard linguistic structures, and add some amount of extra knowledge to the probe. Most recently, \citet{WuChenEtAl_2020_Perturbed_Masking_Parameter-free_Probing_for_Analyzing_and_Interpreting_BERT} proposed a parameter-free approach based on measuring the impact that one word has on predicting another word within a sequence in the MLM task (\autoref{fig:wu-syntax}). They concluded that \textbf{BERT "naturally" learns some syntactic information, although it is not very similar to linguistic annotated resources}. 

The fill-in-the-gap probes of MLM showed that \textbf{BERT takes subject-predicate agreement into account when performing the cloze task} \cite{goldberg2019assessing,vanSchijndel2019quantity}, even for meaningless sentences and sentences with distractor clauses between the subject and the verb \cite{goldberg2019assessing}. 
A study of negative polarity items (NPIs) by \citet{warstadt2019investigating} showed that \textbf{BERT is better able to detect the presence of NPIs} (e.g. "ever") \textbf{and the words that allow their use} (e.g. "whether") \textbf{than scope violations.}

The above claims of syntactic knowledge are belied by the evidence that \textbf{BERT does not "understand" negation and is insensitive to malformed input}. In particular, its predictions were not altered\footnote{See also the recent findings on adversarial triggers, which get the model to produce a certain output even though they are not well-formed from the point of view of a human reader \cite{WallaceFengEtAl_2019_Universal_Adversarial_Triggers_for_Attacking_and_Analyzing_NLP}.} even with shuffled word order, truncated sentences, removed subjects and objects \cite{Ettinger_2019_What_BERT_is_not_Lessons_from_new_suite_of_psycholinguistic_diagnostics_for_language_models}.  This could mean that \textbf{either BERT's syntactic knowledge is incomplete, or it does not need to rely on it for solving its tasks}. The latter seems more likely, since \citet{GlavasVulic_2020_Is_Supervised_Syntactic_Parsing_Beneficial_for_Language_Understanding_Empirical_Investigation} report that an intermediate fine-tuning step with supervised parsing does not make much difference for downstream task performance.

\subsection{Semantic knowledge}

To date, more studies have been devoted to BERT's knowledge of syntactic rather than semantic phenomena. However, we do have evidence from an MLM probing study that \textbf{BERT has some knowledge of semantic roles} 
\cite{Ettinger_2019_What_BERT_is_not_Lessons_from_new_suite_of_psycholinguistic_diagnostics_for_language_models}. BERT even displays some preference for the incorrect fillers for semantic roles that are semantically related to the correct ones, as opposed to those that are unrelated (e.g. "to tip a chef" is better than "to tip a robin", but worse than "to tip a waiter").

\citet{TenneyXiaEtAl_2019_What_do_you_learn_from_context_Probing_for_sentence_structure_in_contextualized_word_representations} showed that \textbf{BERT encodes information about entity types, relations, semantic roles, and proto-roles}, since this information can be detected with probing classifiers. 

\textbf{BERT struggles with representations of numbers.} Addition and number decoding tasks showed that BERT does not form good representations for floating point numbers and fails to generalize away from the training data \cite{wallace2019nlp}. A part of the problem is BERT's wordpiece tokenization, since numbers of similar values can be divided up into substantially different word chunks. 

Out-of-the-box \textbf{BERT is surprisingly brittle to named entity replacements}: e.g. replacing names in the coreference task changes 85\% of predictions \cite{BalasubramanianJainEtAl_2020_Whats_in_Name_Are_BERT_Named_Entity_Representations_just_as_Good_for_any_other_Name}. This suggests that the model does not actually form a generic idea of named entities, although its F1 scores on NER probing tasks are high \cite{tenney2019bert}. \citet{Broscheit_2019_Investigating_Entity_Knowledge_in_BERT_with_Simple_Neural_End-To-End_Entity_Linkinga} find that fine-tuning BERT on Wikipedia entity linking "teaches" it additional entity knowledge, which would suggest that it did not absorb all the relevant entity information during pre-training on Wikipedia.

\subsection{World knowledge}
\label{sec:world-knowledge}

The bulk of evidence about commonsense knowledge captured in BERT comes from practitioners using it to extract such knowledge.  
One direct probing study of BERT reports that \textbf{BERT struggles with pragmatic inference and role-based event knowledge} \cite{Ettinger_2019_What_BERT_is_not_Lessons_from_new_suite_of_psycholinguistic_diagnostics_for_language_models}. BERT also struggles with abstract attributes of objects, as well as visual and perceptual properties that are likely to be assumed rather than mentioned \cite{da2019cracking}.

\begin{figure}[!t]
    \centering
    \includegraphics[width=\linewidth]{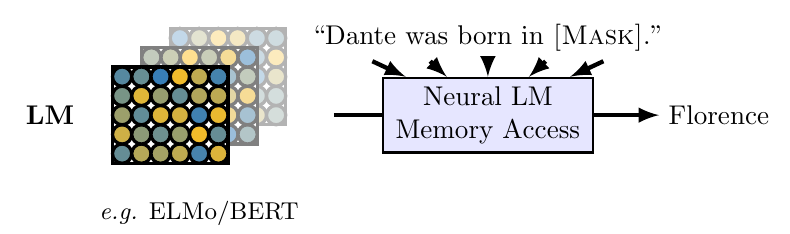}
    \caption{BERT world knowledge ~\protect\cite{PetroniRocktaschelEtAl_2019_Language_Models_as_Knowledge_Bases}}
    \label{fig:bert-kb}
\end{figure}

The MLM component of BERT is easy to adapt for knowledge induction by filling in the blanks (e.g. "Cats like to chase [\textunderscore \textunderscore \textunderscore]"). 
\citet{PetroniRocktaschelEtAl_2019_Language_Models_as_Knowledge_Bases} showed that, \textbf{for some relation types, vanilla BERT is competitive with methods relying on knowledge bases} (\autoref{fig:bert-kb}), and \citet{roberts2020much} show the same for open-domain QA using T5 model \cite{model:t5}. \citet{DavisonFeldmanEtAl_2019_Commonsense_Knowledge_Mining_from_Pretrained_Models} suggest that it generalizes better to unseen data. In order to retrieve BERT's knowledge, we need good template sentences, and there is work on their automatic extraction and augmentation \cite{BouraouiCamacho-ColladosEtAl_2019_Inducing_Relational_Knowledge_from_BERT,JiangXuEtAl_2019_How_Can_We_Know_What_Language_Models_Know}.

However, \textbf{BERT cannot reason based on its world knowledge}.  \citet{ForbesHoltzmanEtAl_Do_Neural_Language_Representations_Learn_Physical_Commonsense} show that BERT can "guess" the affordances and properties of many objects, but can not reason about the relationship between properties and affordances. For example, it ``knows" that people can walk into houses, and that houses are big, but it cannot infer that houses are bigger than people.  \citet{ZhouZhangEtAl_2020_Evaluating_Commonsense_in_Pre-trained_Language_Models} and \citet{RichardsonSabharwal_2019_What_Does_My_QA_Model_Know_Devising_Controlled_Probes_using_Expert_Knowledge} also show that the performance drops with the number of necessary inference steps. Some of BERT's world knowledge success comes from learning stereotypical associations \cite{poerner2019bert}, e.g., a person with an Italian-sounding name is predicted to be Italian, even when it is incorrect. 

\begin{figure*}
    \centering
    \includegraphics[width=.8\linewidth]{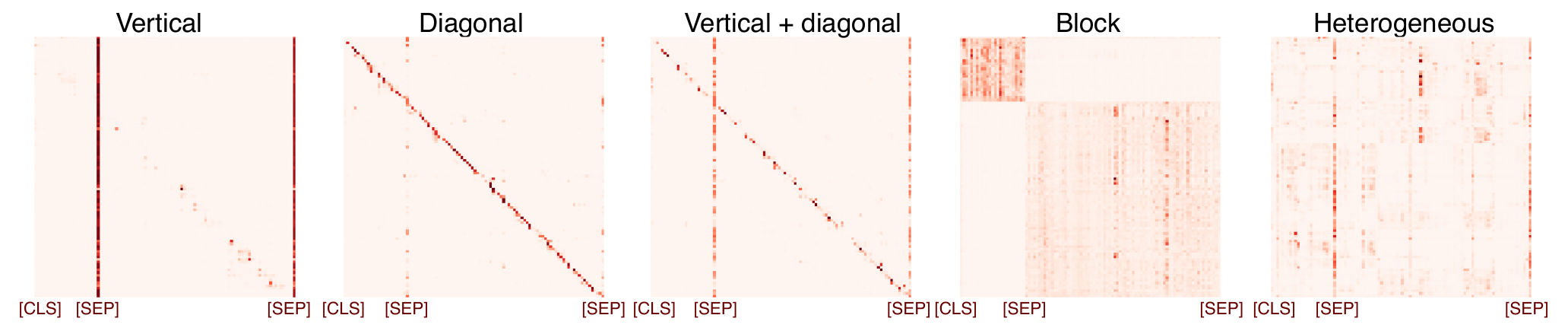}
    \caption{Attention patterns in BERT~\protect\cite{KovalevaRomanovEtAl_2019_Revealing_Dark_Secrets_of_BERT}}
    \label{fig:bert-attn}
\end{figure*}

\subsection{Limitations}
\label{sec:probing-limitations}

Multiple probing studies in \autoref{sec:knowledge} and \autoref{sec:localizing} report that BERT possesses  
a surprising amount of 
syntactic, semantic, and world 
knowledge. However, \citet{tenney2019bert} remarks, ``the fact that a linguistic pattern is not observed by our probing classifier does not guarantee that it is not there, and the observation of a pattern does not tell us how it is used." There is also the issue of how complex a probe should be allowed to be~\cite{liu2019linguistic}. If a more complex probe recovers more information, to what extent are we still relying on the original model?

Furthermore, different probing methods may lead to complementary or even contradictory conclusions, which makes a single test (as in most studies) insufficient \cite{warstadt2019investigating}. A given method might also favor one model over another, e.g., RoBERTa trails BERT with one tree extraction method, but leads with another \cite{htut2019attention}. The choice of linguistic formalism also matters \cite{KuznetsovGurevych_2020_Matter_of_Framing_Impact_of_Linguistic_Formalism_on_Probing_Results}.

In view of all that, the alternative is to focus on identifying what BERT actually relies on at inference time. This direction is currently pursued both at the level of architecture blocks (to be discussed in detail in \autoref{sec:lottery}), and at the level of information encoded in model weights. Amnesic probing \cite{ElazarRavfogelEtAl_2020_When_Bert_Forgets_How_To_POS_Amnesic_Probing_of_Linguistic_Properties_and_MLM_Predictions} aims to specifically remove certain information from the model and see how it changes performance, finding, for example, that language modeling does rely on part-of-speech information.

Another direction is information-theoretic probing. \citet{PimentelValvodaEtAl_2020_Information-Theoretic_Probing_for_Linguistic_Structure} operationalize probing as estimating mutual information between the learned representation and a given linguistic property, which highlights that the focus should be not on the amount of information contained in a representation, but rather on 
how easily it can be extracted from it. \citet{VoitaTitov_2020_Information-Theoretic_Probing_with_Minimum_Description_Length} quantify the amount of effort needed to extract information from a given representation as minimum description length needed to communicate both the probe size and the amount of data required for it to do well on a task.

\section{Localizing linguistic knowledge} 
\label{sec:localizing}

\subsection{BERT embeddings}
\label{sec:embeddings}

In studies of BERT, the term "embedding" refers to the output of a Transformer layer (typically, the final one). Both conventional static embeddings \cite{MikolovSutskeverEtAl_2013_Distributed_representations_of_words_and_phrases_and_their_compositionality} and BERT-style embeddings can be viewed in terms of mutual information maximization \cite{kong2019mutual}, but the latter are \textbf{contextualized}. Every token is represented by a vector dependent on the particular context of occurrence, and contains at least some information about that context \cite{MiaschiDellOrletta_2020_Contextual_and_Non-Contextual_Word_Embeddings_in-depth_Linguistic_Investigation}.

Several studies reported that \textbf{distilled contextualized embeddings better encode lexical semantic information} (i.e. they are better at traditional word-level tasks such as word similarity). The methods to distill a contextualized representation into static include aggregating the information across multiple contexts \cite{AkbikBergmannEtAl_2019_Pooled_Contextualized_Embeddings_for_Named_Entity_Recognition,BommasaniDavisEtAl_2020_Interpreting_Pretrained_Contextualized_Representations_via_Reductions_to_Static_Embeddings}, encoding "semantically bleached" sentences that rely almost exclusively on the meaning of a given word (e.g. "This is <>") \cite{MayWangEtAl_2019_On_Measuring_Social_Biases_in_Sentence_Encoders}, and even using contextualized embeddings to train static embeddings \cite{WangCuiEtAl_2020_How_Can_BERT_Help_Lexical_Semantics_Tasks}.

But this is not to say that there is no room for improvement. \citet{ethayarajh2019contextual} measure how similar the embeddings for identical words are in every layer, reporting that later BERT layers produce more context-specific representations\footnote{\citet{voita2019bottom} look at the evolution of token embeddings, showing that in the earlier Transformer layers, MLM forces the acquisition of contextual information at the expense of the token identity, which gets recreated in later layers.}. 
They also find that BERT embeddings occupy a narrow cone in the vector space, and this effect increases from the earlier to later layers.  That is, \textbf{two random words will on average have a much higher cosine similarity than expected if embeddings were directionally uniform (isotropic)}. Since isotropy was shown to be beneficial for static word embeddings \cite{mu2018allbutthetop}, this might be a fruitful direction to explore for BERT.

Since BERT embeddings are contextualized, an interesting question is to what extent they capture phenomena like polysemy and homonymy. There is indeed evidence that \textbf{BERT's contextualized embeddings form distinct clusters corresponding to word senses} \cite{wiedemann2019does, schmidt2020bert}, making BERT successful at word sense disambiguation task. 
However, \citet{mickus2019you} note that \textbf{the representations of the same word depend on the position of the sentence in which it occurs}, likely due to the NSP objective. This is not desirable from the linguistic point of view, and could be a promising avenue for future work.

The above discussion concerns token embeddings, but BERT is typically used as a sentence or text encoder. The standard way to generate sentence or text representations for classification is to use the [CLS] token, but alternatives are also being discussed, including concatenation of token representations \cite{TanakaShinnouEtAl_2020_Document_Classification_by_Word_Embeddings_of_BERT}, normalized mean \cite{TanakaShinnouEtAl_2020_Document_Classification_by_Word_Embeddings_of_BERT}, and layer activations \cite{MaWangEtAl_2019_Universal_Text_Representation_from_BERT_Empirical_Study}. See \citet{ToshniwalShiEtAl_2020_Cross-Task_Analysis_of_Text_Span_Representations} for a systematic comparison of several methods across tasks and sentence encoders.

\subsection{Self-attention heads}

Several studies proposed classification of attention head types. \citet{RaganatoTiedemann_2018_Analysis_of_Encoder_Representations_in_Transformer-Based_Machine_Translation} discuss attending to the token itself, previous/next tokens and the sentence end. \citet{clark2019does} distinguish between attending to previous/next tokens, \texttt{[CLS]}, \texttt{[SEP]}, punctuation, and "attending broadly" over the sequence. \citet{KovalevaRomanovEtAl_2019_Revealing_Dark_Secrets_of_BERT} propose 5 patterns shown in \autoref{fig:bert-attn}. 

\subsubsection{Heads with linguistic functions}

The "heterogeneous" attention pattern shown in \autoref{fig:bert-attn} \textit{could} potentially be linguistically interpretable, and a number of studies focused on identifying the functions of self-attention heads. In particular, \textbf{some BERT heads seem to specialize in certain types of syntactic relations.} \citet{htut2019attention} and \citet{clark2019does} report that there are BERT heads that attended significantly more than a random baseline to words in certain syntactic positions. The datasets and methods used in these studies differ, but they both find that there are heads that attend to words in \texttt{obj} role more than the positional baseline. The evidence for \texttt{nsubj}, \texttt{advmod}, and \texttt{amod} varies between these two studies. The overall conclusion is also supported by \citet{voita2019analyzing}'s study of the base Transformer in machine translation context.  \citet{HooverStrobeltEtAl_2019_exBERT_Visual_Analysis_Tool_to_Explore_Learned_Representations_in_Transformers_Models} hypothesize that even complex dependencies like \texttt{dobj} are encoded by a combination of heads rather than a single head, but this work is limited to qualitative analysis. \citet{ZhaoBethard_2020_How_does_BERTs_attention_change_when_you_fine-tune_analysis_methodology_and_case_study_in_negation_scopea} looked specifically for the heads encoding negation scope.

Both \citet{clark2019does} and \citet{htut2019attention} conclude that \textbf{no single head has the complete syntactic tree information}, in line with evidence of partial knowledge of syntax (cf. \autoref{sec:syntax}). However, \citet{clark2019does} identify a BERT head that can be directly used as a classifier to perform coreference resolution on par with a rule-based system, which by itself would seem to require quite a lot of syntactic knowledge. 

\citet{lin2019open} present evidence that \textbf{attention weights are weak indicators of subject-verb agreement and reflexive anaphora.} Instead of serving as strong pointers between tokens that should be related, BERT's self-attention weights were close to a uniform attention baseline, but there was some sensitivity to different types of distractors coherent with psycholinguistic data. %
This is consistent with conclusions by \citet{Ettinger_2019_What_BERT_is_not_Lessons_from_new_suite_of_psycholinguistic_diagnostics_for_language_models}.

To our knowledge, morphological information in BERT heads has not been addressed, but with the sparse attention variant by \citet{CorreiaNiculaeEtAl_2019_Adaptively_Sparse_Transformers} in the base Transformer, some attention heads appear to merge BPE-tokenized words. For semantic relations, there are reports of self-attention heads encoding core frame-semantic relations \cite{KovalevaRomanovEtAl_2019_Revealing_Dark_Secrets_of_BERT}, as well as lexicographic and commonsense relations \cite{CuiChengEtAl_2020_Does_BERT_Solve_Commonsense_Task_via_Commonsense_Knowledge}.

The overall popularity of self-attention as an interpretability mechanism is due to the idea that "attention weight has a clear meaning: how much a particular word will be weighted when computing the next representation for the current word" \cite{clark2019does}. This view is currently debated \cite{JainWallace_2019_Attention_is_not_Explanation,SerranoSmith_2019_Is_Attention_Interpretable,WiegreffePinter_2019_Attention_is_not_not_Explanation,BrunnerLiuEtAl_2019_On_Identifiability_in_Transformers}, and in a multi-layer model where attention is followed by non-linear transformations, the patterns in individual heads do not provide a full picture. Also, while many current papers are accompanied by attention visualizations, and there is a growing number of visualization tools \cite{Vig_2019_Visualizing_Attention_in_Transformer-Based_Language_Representation_Models, HooverStrobeltEtAl_2019_exBERT_Visual_Analysis_Tool_to_Explore_Learned_Representations_in_Transformers_Models}, the visualization is typically limited to qualitative analysis (often with cherry-picked examples) \cite{BelinkovGlass_2019_Analysis_Methods_in_Neural_Language_Processing_Survey}, and should not be interpreted as definitive evidence.

\subsubsection{Attention to special tokens}

\citet{KovalevaRomanovEtAl_2019_Revealing_Dark_Secrets_of_BERT} show that \textbf{most self-attention heads do not directly encode any non-trivial linguistic information}, at least when fine-tuned on GLUE  \cite{WangSinghEtAl_2018_GLUE_A_Multi-Task_Benchmark_and_Analysis_Platform_for_Natural_Language_Understanding}, since only less than 50\% of heads exhibit the "heterogeneous" pattern. Much of the model produced the vertical pattern (attention to \texttt{[CLS]}, \texttt{[SEP]}, and punctuation tokens), consistent with the observations by \citet{clark2019does}. This redundancy is likely related to the overparameterization issue (see \autoref{sec:size}).

More recently, \citet{KobayashiKuribayashiEtAl_2020_Attention_Module_is_Not_Only_Weight_Analyzing_Transformers_with_Vector_Norms} showed that the norms of attention-weighted input vectors, which yield a more intuitive interpretation of self-attention, reduce the attention to special tokens. However, even when the attention weights are normed, it is still not the case that most heads that do the "heavy lifting" are even potentially interpretable \cite{PrasannaRogersEtAl_2020_When_BERT_Plays_Lottery_All_Tickets_Are_Winning}. 

One methodological choice in in many studies of attention is to focus on inter-word attention and simply exclude special tokens (e.g. \citet{lin2019open} and \citet{htut2019attention}). However, if attention to special tokens actually matters at inference time, drawing conclusions purely from inter-word attention patterns does not seem warranted. 

The functions of special tokens are not yet well understood. \texttt{[CLS]} is typically viewed as an aggregated sentence-level representation (although all token representations also contain at least some sentence-level information, as discussed in \autoref{sec:embeddings}); in that case, we may not see e.g. full syntactic trees in inter-word attention because part of that information is actually packed in \texttt{[CLS]}. 

    \citet{clark2019does} experiment with encoding Wikipedia paragraphs with base BERT to consider specifically the attention to special tokens, noting that heads in early layers attend more to \texttt{[CLS]}, in middle layers to \texttt{[SEP]}, and in final layers to periods and commas. They hypothesize that its function might be one of "no-op", a signal to ignore the head if its pattern is not applicable to the current case. As a result, for example,  \texttt{[SEP]} gets increased attention starting in layer 5, but its importance for prediction drops. However, after fine-tuning both \texttt{[SEP]} and \texttt{[CLS]} get a lot of attention, depending on the task \cite{KovalevaRomanovEtAl_2019_Revealing_Dark_Secrets_of_BERT}. Interestingly, BERT also pays a lot of attention to punctuation, which \citet{clark2019does} explain by the fact that periods and commas are simply almost as frequent as the special tokens, and so the model might learn to rely on them for the same reasons. 
\subsection{BERT layers}
\label{sec:layers}

The first layer of BERT receives as input a combination of token, segment, and positional embeddings.

It stands to reason that \textbf{the lower layers have the most information about linear word order.} \citet{lin2019open} report a decrease in the knowledge of linear word order around layer 4 in BERT-base. This is accompanied by an increased knowledge of hierarchical sentence structure, as detected by the probing tasks of predicting the token index, the main auxiliary verb and the sentence subject.

\begin{figure}[!t]
	\centering
	\includegraphics[width=.9\linewidth]{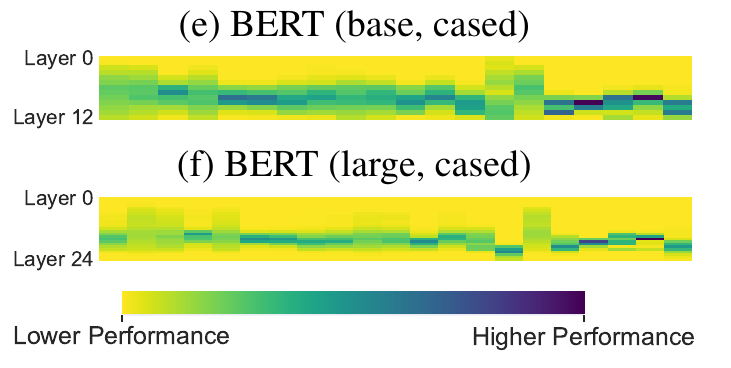}
	\caption{BERT layer transferability (columns correspond to probing tasks, ~\protect\citet{liu2019linguistic}.}
	\label{fig:bert-transferability}
\end{figure}

There is a wide consensus in studies with different tasks, datasets and methodologies that \textbf{syntactic information is most prominent in the middle layers of BERT.\footnote{These BERT results are also compatible with findings by \citet{VigBelinkov_2019_Analyzing_Structure_of_Attention_in_Transformer_Language_Model}, who report the highest attention to tokens in dependency relations in the middle layers of GPT-2.} } \citet{hewitt2019structural} had the most success reconstructing syntactic tree depth from the middle BERT layers (6-9 for base-BERT, 14-19 for BERT-large). \citet{goldberg2019assessing} reports the best subject-verb agreement around layers 8-9, and the performance on syntactic probing tasks used by \citet{jawahar2019does} also seems to peak around the middle of the model.  The prominence of syntactic information in the middle BERT layers is related to \citet{liu2019linguistic}'s observation that the middle layers of Transformers are best-performing overall  and the most transferable across tasks (see \autoref{fig:bert-transferability}). 

There is \textbf{conflicting evidence about syntactic chunks}. \citet{tenney2019bert} conclude that "the basic syntactic information appears earlier in the network while high-level semantic features appear at the higher layers", drawing parallels between this order and the order of components in a typical NLP pipeline -- from POS-tagging to dependency parsing 
to semantic role labeling. \citet{jawahar2019does} also report that the lower layers were more useful for chunking, while middle layers were more useful for parsing. At the same time, the probing experiments by \citet{liu2019linguistic} find the opposite: both POS-tagging and chunking were performed best at the middle layers, in both BERT-base and  BERT-large. However, all three studies use different suites of probing tasks.

\textbf{The final layers of BERT are the most task-specific}. In pre-training, this means specificity to the MLM task, which explains why the middle layers are more transferable \cite{liu2019linguistic}. In fine-tuning, it explains why the final layers change the most \cite{KovalevaRomanovEtAl_2019_Revealing_Dark_Secrets_of_BERT}, and why restoring the weights of lower layers of fine-tuned BERT to their original values does not dramatically hurt the model performance \cite{hao2019visualizing}.

\citet{tenney2019bert} suggest that while syntactic information appears early in the model and can be localized, \textbf{semantics is spread across the entire model}, which explains why certain non-trivial examples get solved incorrectly at first but correctly at the later layers. This is rather to be expected: semantics permeates all language, and linguists debate whether meaningless structures can exist at all \cite[p.166-182]{Goldberg_2006_Constructions_at_Work_The_Nature_of_Generalization_in_Language}. But this raises the question of what stacking more Transformer layers in BERT actually achieves in terms of the spread of semantic knowledge, and whether that is beneficial. \citeauthor{tenney2019bert} compared BERT-base and BERT-large, and found that the overall pattern of cumulative score gains is the same, only more spread out in the larger model. 

Note that \citet{tenney2019bert}'s experiments concern sentence-level semantic relations; \citet{CuiChengEtAl_2020_Does_BERT_Solve_Commonsense_Task_via_Commonsense_Knowledge} report that the encoding of ConceptNet semantic relations is the worst in the early layers and increases towards the top. \citet{jawahar2019does} place "surface features in lower layers, syntactic features in middle layers and semantic features in higher layers", but their conclusion is surprising, given that only one semantic task in this study actually topped at the last layer, and three others peaked around the middle and then considerably degraded by the final layers.

\section{Training BERT}

This section reviews the proposals to optimize the training and architecture of the original BERT.

\subsection{Model architecture choices}

To date, the most systematic study of BERT architecture was performed by \citet{wang2019cross}, who experimented with the number of layers, heads, and model parameters, varying one option and freezing the others. They concluded that \textbf{the number of heads was not as significant as the number of layers}. That is consistent with the findings of \citet{voita2019analyzing} and \citet{michel2019sixteen} (\autoref{sec:size}), and also the observation by \citet{liu2019linguistic} that the middle layers were the most transferable. Larger hidden representation size was consistently better, but the gains varied by setting.

All in all, \textbf{changes in the number of heads and layers appear to perform different functions}. The issue of model depth must be related to the information flow from the most task-specific layers closer to the classifier \cite{liu2019linguistic}, to the initial layers which appear to be the most task-invariant \cite{hao2019visualizing}, and where the tokens resemble the input tokens the most \cite{BrunnerLiuEtAl_2019_On_Identifiability_in_Transformers} (see \autoref{sec:layers}). If that is the case, a deeper model has more capacity to encode information that is not task-specific.

On the other head, many self-attention heads in vanilla BERT seem to naturally learn the same patterns \cite{KovalevaRomanovEtAl_2019_Revealing_Dark_Secrets_of_BERT}. This explains why pruning them does not have too much impact. The question that arises from this is how far we could get with intentionally encouraging diverse self-attention patterns: theoretically, this would mean increasing the amount of information in the model with the same number of weights. \citet{RaganatoScherrerEtAl_2020_Fixed_Encoder_Self-Attention_Patterns_in_Transformer-Based_Machine_Translationa} show for Transformer-based machine translation we can simply pre-set the patterns that we already know the model would learn, instead of learning them from scratch.

Vanilla BERT is symmetric and balanced in terms of self-attention and feed-forward layers, but it may not have to be. For the base Transformer, \citet{PressSmithEtAl_2020_Improving_Transformer_Models_by_Reordering_their_Sublayers} report benefits from more self-attention sublayers at the bottom and more feedforward sublayers at the top.

\subsection{Improvements to the training regime}

\citet{model:roberta} demonstrate \textbf{the benefits of large-batch training}: with 8k examples both the language model perplexity and downstream task performance are improved. They also publish their recommendations for other parameters. \citet{you2019large} report that with a batch size of 32k BERT's training time can be significantly reduced with no degradation in performance. \citet{zhou2019improving} observe that the normalization of the trained \texttt{[CLS]} token stabilizes the training and slightly improves performance on text classification tasks. 

\citet{gong2019efficient} note that, since self-attention patterns in higher and lower layers are similar, \textbf{the model training can be done in a recursive manner}, where the shallower version is trained first and then the trained parameters are copied to deeper layers. Such a "warm-start" can lead to a 25\% faster training without sacrificing performance.

\subsection{Pre-training BERT}

The original BERT is a bidirectional Transformer pre-trained on two tasks: next sentence prediction (NSP) and masked language model (MLM) (\autoref{sec:bert}). Multiple studies have come up with \textbf{alternative training objectives} to improve on BERT, which could be categorized as follows:

\begin{itemize*}

\item \textbf{How to mask.} 
\citet{model:t5} systematically experiment with corruption rate and corrupted span length. \citet{model:roberta} propose diverse masks for training examples within an epoch, while \citet{BaevskiEdunovEtAl_2019_Cloze-driven_Pretraining_of_Self-attention_Networks} mask every token in a sequence instead of a random selection. \citet{ClinchantJungEtAl_2019_On_use_of_BERT_for_Neural_Machine_Translation} replace the MASK token with \texttt{[UNK]} token, to help the model learn a representation for unknowns that could be useful for translation. \citet{SongTanEtAl_2020_MPNet_Masked_and_Permuted_Pre-training_for_Language_Understanding} maximize the amount of information available to the model by conditioning on both masked and unmasked tokens, and letting the model see how many tokens are missing.

\item \textbf{What to mask.} Masks can be applied to full words instead of word-pieces \cite{devlin2019bert,CuiCheEtAl_2019_Pre-Training_with_Whole_Word_Masking_for_Chinese_BERT}. Similarly, we can mask spans rather than single tokens \cite{JoshiChenEtAl_2020_SpanBERT_Improving_Pre-training_by_Representing_and_Predicting_Spans}, predicting how many are missing \cite{model_bart}. Masking phrases and named entities \cite{SunWangEtAl_2019_ERNIE_Enhanced_Representation_through_Knowledge_Integration} improves representation of structured knowledge. 

\item \textbf{Where to mask.} \citet{LampleConneau_2019_Cross-lingual_Language_Model_Pretraining} 
use arbitrary text streams instead of sentence pairs and subsample frequent outputs similar to \citet{MikolovSutskeverEtAl_2013_Distributed_representations_of_words_and_phrases_and_their_compositionality}. \citet{BaoDongEtAl_2020_UniLMv2_Pseudo-Masked_Language_Models_for_Unified_Language_Model_Pre-Training} combine the standard autoencoding MLM with partially autoregressive LM objective using special pseudo mask tokens.

\item \textbf{Alternatives to masking.} \citet{model:t5} experiment with  replacing and dropping spans, \citet{model_bart} explore deletion, infilling, sentence permutation and document rotation, and \citet{model:ernie} predict whether a token is capitalized and whether it occurs in other segments of the same document. 
\citet{model:xlnet} train on different permutations of word order in the input sequence, maximizing the probability of the original word order (cf. the n-gram word order reconstruction task \cite{WangBiEtAl_2019_StructBERT_Incorporating_Language_Structures_into_Pre-training_for_Deep_Language_Understanding}). \citet{ClarkLuongEtAl_2020_ELECTRA_Pre-training_Text_Encoders_as_Discriminators_Rather_Than_Generators} detect tokens that were replaced by a generator network rather than masked. 

\item \textbf{NSP alternatives.} Removing NSP does not hurt or slightly improves performance \cite{model:roberta,JoshiChenEtAl_2020_SpanBERT_Improving_Pre-training_by_Representing_and_Predicting_Spans,ClinchantJungEtAl_2019_On_use_of_BERT_for_Neural_Machine_Translation}.  \citet{WangBiEtAl_2019_StructBERT_Incorporating_Language_Structures_into_Pre-training_for_Deep_Language_Understanding} and \citet{ChengXuEtAl_2019_Symmetric_Regularization_based_BERT_for_Pair-wise_Semantic_Reasoning} replace NSP with the task of predicting both the next and the previous sentences. \citet{model:albert} replace the negative NSP examples by swapped sentences from positive examples, rather than sentences from different documents. ERNIE 2.0 includes sentence reordering and sentence distance prediction. \citet{BaiShiEtAl_2020_SegaBERT_Pre-training_of_Segment-aware_BERT_for_Language_Understanding} replace both NSP and token position embeddings by a combination of paragraph, sentence, and token index embeddings. \citet{LiChoi_2020_Transformers_to_Learn_Hierarchical_Contexts_in_Multiparty_Dialogue_for_Span-based_Question_Answering} experiment with utterance order prediction task for multi-party dialogue (and also MLM at the level of utterances and the whole dialogue).
    
\item \textbf{Other tasks.} \citet{model:ernie} propose simultaneous learning of 7 tasks, including discourse relation classification and predicting whether a segment is relevant for IR. \citet{GuuLeeEtAl_2020_REALM_Retrieval-Augmented_Language_Model_Pre-Training} include a latent knowledge retriever in language model pretraining. \citet{WangGaoEtAl_2020_KEPLER_Unified_Model_for_Knowledge_Embedding_and_Pre-trained_Language_Representation} combine MLM with knowledge base completion objective. \citet{GlassGliozzoEtAl_2020_Span_Selection_Pre-training_for_Question_Answering} replace MLM with span prediction task (as in extractive question answering), where the model is expected to provide the answer not from its own weights, but from a \textit{different} passage containing the correct answer (a relevant search engine query snippet). 
\end{itemize*}    

Another obvious source of improvement is pre-training data. Several studies explored the benefits of increasing the corpus volume \cite{model:roberta, ConneauKhandelwalEtAl_2019_Unsupervised_Cross-lingual_Representation_Learning_at_Scale,BaevskiEdunovEtAl_2019_Cloze-driven_Pretraining_of_Self-attention_Networks} and longer training \cite{model:roberta}. The data also does not have to be raw text: there is a number efforts to \textbf{incorporate explicit linguistic information}, both syntactic \cite{SundararamanSubramanianEtAl_2019_Syntax-Infused_Transformer_and_BERT_models_for_Machine_Translation_and_Natural_Language_Understanding} and semantic \cite{ZhangWuEtAl_2020_Semantics-aware_BERT_for_Language_Understanding}. \citet{WuLvEtAl_2019_Conditional_BERT_Contextual_Augmentation} and \citet{KumarChoudharyEtAl_2020_Data_Augmentation_using_Pre-trained_Transformer_Models} include the label for a given sequence from an annotated task dataset. \citet{SchickSchutze_2020_BERTRAM_Improved_Word_Embeddings_Have_Big_Impact_on_Contextualized_Model_Performance} separately learn representations for rare words. 

Although BERT is already actively used as a source of world knowledge (see \autoref{sec:world-knowledge}), there is also work on \textbf{explicitly supplying structured knowledge}. One approach is entity-enhanced models. For example,  \citet{PetersNeumannEtAl_2019_Knowledge_Enhanced_Contextual_Word_Representations,ZhangHanEtAl_2019_ERNIE_Enhanced_Language_Representation_with_Informative_Entities} include entity embeddings as input for training BERT, while \citet{poerner2019bert} adapt entity vectors to BERT representations. As mentioned above,  \citet{WangGaoEtAl_2020_KEPLER_Unified_Model_for_Knowledge_Embedding_and_Pre-trained_Language_Representation} integrate knowledge not through entity embeddings, but through additional pre-training objective of knowledge base completion. \citet{SunWangEtAl_2019_ERNIE_Enhanced_Representation_through_Knowledge_Integration,model:ernie} modify the standard MLM task to mask named entities rather than random words, and \citet{YinNeubigEtAl_2020_TaBERT_Pretraining_for_Joint_Understanding_of_Textual_and_Tabular_Data} train with MLM objective over both text and linearized table data. \citet{WangTangEtAl_2020_K-Adapter_Infusing_Knowledge_into_Pre-Trained_Models_with_Adapters} enhance RoBERTa with both linguistic and factual knowledge with task-specific adapters.

\begin{figure}[!t]
	\centering
	\includegraphics[width=.9\linewidth]{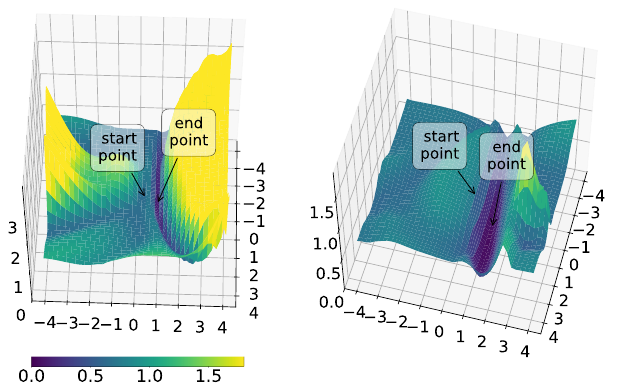}
	\caption{Pre-trained weights help BERT find wider optima in fine-tuning on MRPC (right) than training from scratch (left)~\protect\cite{hao2019visualizing}}
	\label{fig:bert-loss}
\end{figure}

Pre-training is the most expensive part of training BERT, and it would be informative  to know how much benefit it provides. On some tasks, a randomly initialized and fine-tuned BERT obtains competitive or higher results than the pre-trained BERT with the task classifier and frozen weights \cite{KovalevaRomanovEtAl_2019_Revealing_Dark_Secrets_of_BERT}. The consensus in the community is that pre-training does help in most situations, but the degree and its exact contribution requires further investigation. \citet{PrasannaRogersEtAl_2020_When_BERT_Plays_Lottery_All_Tickets_Are_Winning} found that \textit{most} weights of pre-trained BERT are useful in fine-tuning, although there are "better" and "worse" subnetworks. One explanation is that pre-trained weights help the fine-tuned BERT find wider and flatter areas with smaller generalization error, which makes the model more robust to overfitting (see \autoref{fig:bert-loss} from \citet{hao2019visualizing}). 

Given the large number and variety of proposed modifications, one would wish to know how much impact each of them has.  However, due to the overall trend towards large model sizes, systematic ablations have become expensive. Most new models claim superiority on standard benchmarks, but gains are often marginal, and estimates of model stability and significance testing are very rare.

\subsection{Fine-tuning BERT}

Pre-training + fine-tuning workflow is a crucial part of BERT. The former is supposed to provide task-independent knowledge, and the latter would presumably teach the model to rely more on the representations useful for the task at hand. 

\citet{KovalevaRomanovEtAl_2019_Revealing_Dark_Secrets_of_BERT} did not find that to be the case for BERT fine-tuned on GLUE tasks\footnote{\citet{KondratyukStraka_2019_75_Languages_1_Model_Parsing_Universal_Dependencies_Universally} suggest that fine-tuning on Universal Dependencies does result in  syntactically meaningful attention patterns, but there was no quantitative evaluation.}: during fine-tuning, the most changes for 3 epochs occurred in the last two layers of the models,  but those changes caused self-attention to focus on \texttt{[SEP]} rather than on linguistically interpretable patterns.  It is understandable why fine-tuning would increase the attention to \texttt{[CLS]}, but not \texttt{[SEP]}. If \citet{clark2019does} are correct that \texttt{[SEP]} serves as "no-op" indicator, fine-tuning basically tells BERT what to ignore.

Several studies explored the possibilities of improving the fine-tuning of BERT:

\begin{itemize*}
    \item \textbf{Taking more layers into account}: learning a complementary representation of the information in deep and output layers \cite{YangZhao_2019_Deepening_Hidden_Representations_from_Pre-trained_Language_Models_for_Natural_Language_Understanding}, using a weighted combination of all layers instead of the final one \cite{SuCheng_2019_SesameBERT_Attention_for_Anywhere,KondratyukStraka_2019_75_Languages_1_Model_Parsing_Universal_Dependencies_Universally}, and layer dropout \cite{KondratyukStraka_2019_75_Languages_1_Model_Parsing_Universal_Dependencies_Universally}.
    \item \textbf{Two-stage fine-tuning} introduces an intermediate supervised training stage between pre-training and fine-tuning  \cite{PhangFevryEtAl_2019_Sentence_Encoders_on_STILTs_Supplementary_Training_on_Intermediate_Labeled-data_Tasks,GargVuEtAl_2020_TANDA_Transfer_and_Adapt_Pre-Trained_Transformer_Models_for_Answer_Sentence_Selection,AraseTsujii_2019_Transfer_Fine-Tuning_BERT_Case_Study,PruksachatkunPhangEtAl_2020_Intermediate-Task_Transfer_Learning_with_Pretrained_Language_Models_When_and_Why_Does_It_Work,GlavasVulic_2020_Is_Supervised_Syntactic_Parsing_Beneficial_for_Language_Understanding_Empirical_Investigation}. 
    \citet{Ben-DavidRabinovitzEtAl_2020_PERL_Pivot-based_Domain_Adaptation_for_Pre-trained_Deep_Contextualized_Embedding_Models} propose a pivot-based variant of MLM to fine-tune BERT for domain adaptation.
    \item \textbf{Adversarial token perturbations} improve robustness of the model \cite{ZhuChengEtAl_2019_FreeLB_Enhanced_Adversarial_Training_for_Language_Understanding}.
    
    \item \textbf{Adversarial regularization} in combination with \textit{Bregman Proximal Point Optimization} helps alleviate pre-trained knowledge forgetting and therefore prevents BERT from overfitting to downstream tasks \cite{jiang2019smart}.
    
    \item \textbf{Mixout regularization} improves the stability of BERT fine-tuning even
    for a small number of training examples \cite{lee2019mixout}.
\end{itemize*}

With large models, even fine-tuning becomes expensive, but \citet{HoulsbyGiurgiuEtAl_2019_Parameter-Efficient_Transfer_Learning_for_NLP} show that it can be successfully approximated with adapter modules. They achieve competitive performance on 26 classification tasks at a fraction of the computational cost. Adapters in BERT were also used for multi-task learning \cite{SticklandMurray_2019_BERT_and_PALs_Projected_Attention_Layers_for_Efficient_Adaptation_in_Multi-Task_Learning} and cross-lingual transfer \cite{ArtetxeRuderEtAl_2019_On_Cross-lingual_Transferability_of_Monolingual_Representations}. An alternative to fine-tuning is extracting features from frozen representations, but fine-tuning works better for BERT  
 \cite{PetersRuderEtAl_2019_To_Tune_or_Not_to_Tune_Adapting_Pretrained_Representations_to_Diverse_Tasks}.

A big methodological challenge in the current NLP is that the reported performance improvements of new models may well be within variation induced by environment factors \cite{Crane_2018_Questionable_Answers_in_Question_Answering_Research_Reproducibility_and_Variability_of_Published_Results}. BERT is not an exception.  \citet{DodgeIlharcoEtAl_2020_Fine-Tuning_Pretrained_Language_Models_Weight_Initializations_Data_Orders_and_Early_Stopping} report significant variation for BERT fine-tuned on GLUE tasks due to both weight initialization and training data order. They also propose early stopping on the less-promising seeds. 

Although we hope that the above observations may be useful for the practitioners, this section does not exhaust the current research on fine-tuning and its alternatives. For example, we do not cover such topics as Siamese architectures, policy gradient training, automated curriculum learning, and others.

\section{How big should BERT be?}
\label{sec:size}

\begin{table*}[t]
\scriptsize
\begin{tabular}{p{.2cm} p{4.1cm}p{1.3cm}p{1.3cm}p{.9cm}p{.9cm}p{4.3cm}}
\toprule
& & \textbf{Compression} & \textbf{Performance} & \textbf{Speedup} & \textbf{Model} & \textbf{Evaluation} \\ \toprule
&  BERT-base \cite{devlin2019bert} & $\times$1 & 100\% & $\times$1 & BERT\textsubscript{12}& All GLUE tasks, SQuAD \\
&  BERT-small & $\times$3.8 & 91\% & - & BERT\textsubscript{4}$\dagger$ & All GLUE tasks \\ \midrule
\multirow{13}{*}{\rotatebox[origin=c]{90}{Distillation}} & DistilBERT \cite{sanh2019distilbert} & $\times$1.5 & 90\%$^{\S}$ & $\times$1.6 & BERT\textsubscript{6} & All GLUE tasks, SQuAD \\
& BERT\textsubscript{6}-PKD \cite{sun2019patient} & $\times$1.6 & 98\% & $\times$1.9 & BERT\textsubscript{6} & No WNLI, CoLA, STS-B; RACE \\
& BERT\textsubscript{3}-PKD \cite{sun2019patient} & $\times$2.4 & 92\% & $\times$3.7 & BERT\textsubscript{3} & No WNLI, CoLA, STS-B; RACE \\
& \citet{aguilar2019knowledge}, Exp. 3 & $\times$1.6 & 93\% & - & BERT\textsubscript{6} & CoLA, MRPC, QQP, RTE \\
& BERT-48 \cite{zhao2019extreme} & $\times$62 & 87\% & $\times$77 & BERT\textsubscript{12}$^{*\dagger}$ & MNLI, MRPC, SST-2 \\
& BERT-192 \cite{zhao2019extreme} & $\times$5.7 & 93\% & $\times$22 & BERT\textsubscript{12}$^{*\dagger}$ & MNLI, MRPC, SST-2 \\
& TinyBERT \cite{jiao2019tinybert} & $\times$7.5 & 96\% & $\times$9.4 & BERT\textsubscript{4}$^{\dagger}$ & No WNLI; SQuAD\\
& MobileBERT \cite{sunmobilebert} & $\times$4.3 & 100\% & $\times$4 & BERT\textsubscript{24}$^{\dagger}$ & No WNLI; SQuAD \\
& PD \cite{turc2019well} & $\times$1.6 & 98\% & $\times$2.5$^\ddagger$ & BERT\textsubscript{6}$^{\dagger}$ & No WNLI, CoLA and STS-B \\
& WaLDORf \cite{tian2019waldorf} &$\times$4.4 & 93\% & $\times$9& BERT\textsubscript{8}$^{\dagger\|}$ & SQuAD\\
& MiniLM \cite{wang2020minilm} & $\times$1.65 & 99\% & $\times$2 & BERT\textsubscript{6} & No WNLI, STS-B, MNLI\textsubscript{mm}; SQuAD\\
& MiniBERT\cite{tsai2019small} & $\times$6$^{**}$ & 98\% & $\times$27$^{**}$ & mBERT\textsubscript{3}$^{\dagger}$ & CoNLL-18 POS and morphology \\
& BiLSTM-soft \cite{tang2019distilling} & $\times$110 & 91\% & $\times$434$^{\ddagger}$ & BiLSTM\textsubscript{1} & MNLI, QQP, SST-2 \\
 \midrule
\multirow{2}{*}{\rotatebox[origin=c]{90}{\parbox{1cm}{Quanti-\\zation}}} & Q-BERT-MP \cite{shen2019q} & $\times$13 & 98\%$^\P$ & - & BERT\textsubscript{12} & MNLI, SST-2, CoNLL-03, SQuAD \\
& BERT-QAT \cite{zafrir2019q8bert} & $\times$4 & 99\% & - & BERT\textsubscript{12} & No WNLI, MNLI; SQuAD \\
& GOBO\cite{ZadehMoshovos_2020_GOBO_Quantizing_Attention-Based_NLP_Models_for_Low_Latency_and_Energy_Efficient_Inference} & $\times 9.8$ & 99\% & - & BERT\textsubscript{12} & MNLI \\ 
  \midrule
\multirow{4}{*}{\rotatebox[origin=c]{90}{Pruning}} & \citet{McCarleyChakravartiEtAl_2020_Structured_Pruning_of_BERT-based_Question_Answering_Model}, ff2 & $\times$2.2$^\ddagger$ & 98\%$^\ddagger$ & $\times$1.9$^\ddagger$ & BERT\textsubscript{24} & SQuAD, Natural Questions \\
& RPP \cite{GuoLiuEtAl_2019_Reweighted_Proximal_Pruning_for_Large-Scale_Language_Representation} & $\times$1.7$^{\ddagger}$  & 99\%$^{\ddagger}$& - & BERT\textsubscript{24} & No WNLI, STS-B; SQuAD \\
& Soft MvP \cite{SanhWolfEtAl_2020_Movement_Pruning_Adaptive_Sparsity_by_Fine-Tuning} & $\times$33 & 94\%$^\P$ & - & BERT\textsubscript{12} & MNLI, QQP, SQuAD \\
&IMP \cite{ChenFrankleEtAl_2020_Lottery_Ticket_Hypothesis_for_Pre-trained_BERT_Networks}, rewind 50\% & $\times$1.4--2.5 & 94--100\% & - & BERT\textsubscript{12} & No MNLI-mm; SQuAD \\
\midrule
 
\multirow{4}{*}{\rotatebox[origin=c]{90}{Other}} & ALBERT-base \cite{lan2019albert} & $\times$9 & 97\% & - & BERT\textsubscript{12}$^{\dagger}$ & MNLI, SST-2 \\
 & ALBERT-xxlarge \cite{lan2019albert} & $\times$0.47 & 107\% & - & BERT\textsubscript{12}$^{\dagger}$ & MNLI, SST-2 \\
 & BERT-of-Theseus \cite{xu2020bert} & $\times$1.6 & 98\% & $\times$1.9 & BERT\textsubscript{6} & No WNLI \\
 & PoWER-BERT \cite{goyal2020power}& N/A & 99\% & $\times$2--4.5 & BERT\textsubscript{12} & No WNLI; RACE \\
\bottomrule

\end{tabular}

\caption[Caption]{Comparison of BERT compression studies. Compression, performance retention, inference time speedup figures are given with respect to BERT\textsubscript{base}, unless indicated otherwise. Performance retention is measured as a ratio of average scores achieved by a given model and by BERT\textsubscript{base}. The subscript in the model description reflects the number of layers used.  $^{*}$Smaller vocabulary used. $^{\dagger}$The dimensionality of the hidden layers is reduced. $^{\|}$Convolutional layers used. $^{\ddagger}$Compared to BERT\textsubscript{large}. $^{**}$Compared to mBERT. $^\S$As reported in \cite{jiao2019tinybert}.$^\P$In comparison to the dev set.}
\label{tab:compression}
\end{table*}

\subsection{Overparameterization}

Transformer-based models keep growing by orders of magnitude: the 110M parameters of base BERT are now dwarfed by 17B parameters of Turing-NLG \cite{turing-nlg}, which is dwarfed by 175B of GPT-3 \cite{BrownMannEtAl_2020_Language_Models_are_Few-Shot_Learners}. This trend raises concerns about computational complexity of self-attention \cite{WuFanEtAl_2019_Pay_Less_Attention_with_Lightweight_and_Dynamic_Convolutions}, environmental issues \cite{StrubellGaneshEtAl_2019_Energy_and_Policy_Considerations_for_Deep_Learning_in_NLP,SchwartzDodgeEtAl_2019_Green_AI}, fair comparison of architectures \cite{AssenmacherHeumann_2020_On_comparability_of_Pre-trained_Language_Models}, and reproducibility. 

Human language is incredibly complex, and would perhaps take many more parameters to describe fully, but the current models do not make good use of the parameters they already have. \citet{voita2019analyzing} showed that \textbf{all but a few Transformer heads could be pruned without significant losses in performance}. For BERT, \citet{clark2019does} observe that most heads in the same layer show similar self-attention patterns (perhaps related to the fact that the output of all self-attention heads in a layer is passed through the same MLP), which explains why \citet{michel2019sixteen} were able to reduce most layers to a single head.

Depending on the task, some BERT heads/layers are not only redundant \cite{kao2020further}, but also harmful to the downstream task performance. \textbf{Positive effect from head disabling} was reported for machine translation \cite{michel2019sixteen}, abstractive summarization \cite{baan2019understanding}, and GLUE tasks \cite{KovalevaRomanovEtAl_2019_Revealing_Dark_Secrets_of_BERT}. Additionally, \citet{tenney2019bert} examine the cumulative gains of their structural probing classifier, observing that in 5 out of 8 probing tasks some layers cause a drop in scores (typically in the final layers). \citet{gordon2020compressing} find that 30--40\% of the weights can be pruned without impact on downstream tasks.

In general, larger BERT models perform better \cite{liu2019linguistic,roberts2020much}, but not always: BERT-base outperformed BERT-large on subject-verb agreement \cite{goldberg2019assessing} and sentence subject detection \cite{lin2019open}. 
Given the complexity of language, and amounts of pre-training data, 
it is not clear why BERT ends up with redundant heads and layers. \citet{clark2019does} suggest that one possible reason is the use of attention dropouts, which causes some attention weights to be zeroed-out during training.

\subsection{Compression techniques}

Given the above evidence of overparameterization, it does not come as a surprise that \textbf{BERT can be efficiently compressed with minimal accuracy loss}, which would be highly desirable for real-world applications. Such efforts to date are summarized in \autoref{tab:compression}. The main approaches are knowledge distillation, quantization, and pruning.

The studies in the \textbf{knowledge distillation framework} \cite{hinton2015distilling} use a smaller student-network trained to mimic the behavior of a larger teacher-network. For BERT, this has been achieved through experiments with loss functions \cite{model:distilBERT,jiao2019tinybert}, mimicking the activation patterns of individual portions of the teacher network  \cite{sun2019patient}, and knowledge transfer at the pre-training \cite{turc2019well, jiao2019tinybert, sunmobilebert} or fine-tuning stage \cite{jiao2019tinybert}. \citet{McCarleyChakravartiEtAl_2020_Structured_Pruning_of_BERT-based_Question_Answering_Model} suggest that distillation has so far worked better for GLUE than for reading comprehension, and report good results for QA from a combination of structured pruning and task-specific distillation. 

\textbf{Quantization} decreases BERT's memory footprint through lowering the precision of its weights \cite{shen2019q, zafrir2019q8bert}. Note that this strategy often requires compatible hardware.

As discussed in \autoref{sec:size}, individual self-attention heads and BERT layers can be disabled without significant drop in performance \cite{michel2019sixteen,KovalevaRomanovEtAl_2019_Revealing_Dark_Secrets_of_BERT,baan2019understanding}. \textbf{Pruning} is a compression technique that takes advantage of that fact, typically reducing the amount of computation via zeroing out of certain parts of the large model. In structured pruning, architecture blocks are dropped, as in LayerDrop \cite{FanGraveEtAl_2019_Reducing_Transformer_Depth_on_Demand_with_Structured_Dropout}. In unstructured, the weights in the entire model are pruned irrespective of their location, as in magnitude pruning \cite{ChenFrankleEtAl_2020_Lottery_Ticket_Hypothesis_for_Pre-trained_BERT_Networks} or movement pruning \cite{SanhWolfEtAl_2020_Movement_Pruning_Adaptive_Sparsity_by_Fine-Tuning}. 

\citet{PrasannaRogersEtAl_2020_When_BERT_Plays_Lottery_All_Tickets_Are_Winning} and 
\citet{ChenFrankleEtAl_2020_Lottery_Ticket_Hypothesis_for_Pre-trained_BERT_Networks} explore BERT from the perspective of the lottery ticket hypothesis \cite{FrankleCarbin_2019_Lottery_Ticket_Hypothesis_Finding_Sparse_Trainable_Neural_Networks}, looking specifically at the "winning" subnetworks in pre-trained BERT. They independently find that such subnetworks do exist, and that transferability between subnetworks for different tasks varies.

If the ultimate goal of training BERT is compression, \citet{li2020train} recommend training larger models and compressing them heavily rather than compressing smaller models lightly. 

Other techniques include decomposing BERT's embedding matrix into smaller matrices \cite{model:albert}, progressive module replacing \cite{xu2020bert} and dynamic elimination of intermediate encoder outputs \cite{goyal2020power}. 
See \citet{ganesh2020compressing} for a more detailed discussion of compression methods. 

\subsection{Pruning and model analysis}
\label{sec:lottery}

There is a nascent discussion around pruning as a model analysis technique. The basic idea is that a compressed model a priori consists of elements that are useful for prediction; therefore by finding out what they do we may find out what the whole network does. For instance, BERT has heads that seem to encode frame-semantic relations, but disabling them might not hurt downstream task performance \citet{KovalevaRomanovEtAl_2019_Revealing_Dark_Secrets_of_BERT}; this suggests that this knowledge is not actually used. 

For the base Transformer, \citet{voita2019analyzing} identify the functions of self-attention heads and then check which of them survive the pruning, finding that the syntactic and positional heads are the last ones to go. For BERT, \citet{PrasannaRogersEtAl_2020_When_BERT_Plays_Lottery_All_Tickets_Are_Winning} go in the opposite direction: pruning on the basis of importance scores, and interpreting the remaining "good" subnetwork. With respect to self-attention heads specifically, it does not seem to be the case that only the heads that potentially encode non-trivial linguistic patterns survive the pruning. 

The models and methodology in these studies differ, so the evidence is inconclusive. In particular, \citet{voita2019analyzing} find that before pruning the majority of heads are syntactic, and \citet{PrasannaRogersEtAl_2020_When_BERT_Plays_Lottery_All_Tickets_Are_Winning} -- that the majority of heads do not have potentially non-trivial attention patterns.

An important limitation of the current head and layer ablation studies
\cite{michel2019sixteen,KovalevaRomanovEtAl_2019_Revealing_Dark_Secrets_of_BERT} is that they inherently assume that certain knowledge is contained in heads/layers. However, there is evidence of more diffuse representations spread across the full network, such as the gradual increase in accuracy on difficult semantic parsing tasks \cite{tenney2019bert} or the absence of heads that would perform parsing "in general" \cite{clark2019does,htut2019attention}. If so, ablating individual components harms the weight-sharing mechanism. Conclusions from component ablations are also problematic if the same information is duplicated elsewhere in the network.

\section{Directions for further research}

BERTology has clearly come a long way, but it is fair to say we still have more questions than answers about how BERT works. In this section, we list what we believe to be the most promising directions for further research.

\textbf{Benchmarks that require verbal reasoning.} While BERT enabled breakthroughs on many NLP benchmarks, a growing list of analysis papers are showing that its language skills are not as impressive as it seems. In particular, it was shown to rely on shallow heuristics in natural language inference \cite{mccoy2019right,zellers2019hellaswag,JinJinEtAl_2020_Is_BERT_Really_Robust_Strong_Baseline_for_Natural_Language_Attack_on_Text_Classification_and_Entailment}, reading comprehension \cite{si2019does,RogersKovalevaEtAl_2020_Getting_Closer_to_AI_Complete_Question_Answering_Set_of_Prerequisite_Real_Tasks,SugawaraStenetorpEtAl_2020_Assessing_Benchmarking_Capacity_of_Machine_Reading_Comprehension_Datasets,SiWangEtAl_2019_What_does_BERT_Learn_from_Multiple-Choice_Reading_Comprehension_Datasets,YogatamadAutumeEtAl_2019_Learning_and_Evaluating_General_Linguistic_Intelligencea}, argument reasoning comprehension \cite{NivenKao_2019_Probing_Neural_Network_Comprehension_of_Natural_Language_Argumentsa}, and text classification \cite{JinJinEtAl_2020_Is_BERT_Really_Robust_Strong_Baseline_for_Natural_Language_Attack_on_Text_Classification_and_Entailment}. Such heuristics can even be used to reconstruct a non-publicly-available model \cite{KrishnaTomarEtAl_2020_Thieves_on_Sesame_Street_Model_Extraction_of_BERT-based_APIs}. As with any optimization method, if there is a shortcut in the data, we have no reason to expect BERT to not learn it. But harder datasets that cannot be resolved with shallow heuristics are unlikely to emerge if their development is not as valued as modeling work. 

\textbf{Benchmarks for the full range of linguistic competence.} While the language models seem to acquire a great deal of knowledge about language, we do not currently have comprehensive stress tests for different aspects of linguistic knowledge. A step in this direction is the "Checklist" behavioral testing \cite{RibeiroWuEtAl_2020_Beyond_Accuracy_Behavioral_Testing_of_NLP_Models_with_CheckList}, the best paper at ACL 2020. Ideally, such tests would measure not only errors, but also sensitivity \cite{Ettinger_2019_What_BERT_is_not_Lessons_from_new_suite_of_psycholinguistic_diagnostics_for_language_models}.

\paragraph{Developing methods to "teach" reasoning.} While large pre-trained models have a lot of knowledge, they often fail if any reasoning needs to be performed on top of the facts they possess \cite[see also \autoref{sec:world-knowledge}]{TalmorElazarEtAl_2019_oLMpics_-_On_what_Language_Model_Pre-training_Captures}. For instance, \citet{richardson2019probing} propose a method to "teach" BERT quantification, conditionals, comparatives, and boolean coordination. 

\paragraph{Learning what happens at inference time.} Most BERT analysis papers focus on different probes of the model, with the goal to find what the language model "knows". However, probing studies have limitations (\autoref{sec:probing-limitations}), and to this point, far fewer papers have focused on discovering what knowledge actually gets used. Several promising directions are the "amnesic probing" \cite{ElazarRavfogelEtAl_2020_When_Bert_Forgets_How_To_POS_Amnesic_Probing_of_Linguistic_Properties_and_MLM_Predictions}, identifying features important for prediction for a given task \cite{arkhangelskaia2019whatcha}, and pruning the model to remove the non-important components \cite{voita2019analyzing,michel2019sixteen,PrasannaRogersEtAl_2020_When_BERT_Plays_Lottery_All_Tickets_Are_Winning}. 

\section{Conclusion}

In a little over a year, BERT has become a ubiquitous baseline in NLP experiments and inspired numerous studies analyzing the model and proposing various improvements. The stream of papers seems to be accelerating rather than slowing down, and we hope that this survey helps the community to focus on the biggest unresolved questions.

\section{Acknowledgements}

We thank the anonymous reviewers for their valuable feedback. This work is funded in part by the NSF award number IIS-1844740 to Anna Rumshisky.

\bibliography{tacl2018}
\bibliographystyle{acl_natbib}

\end{document}